\documentclass[10pt,twocolumn,letterpaper]{article}
\usepackage[pagenumbers]{iccv}
\usepackage{paralist}
\usepackage{enumitem}
\usepackage{multirow}
\usepackage{titletoc}
\usepackage{booktabs}
\usepackage{gensymb}
\usepackage{bm}
\usepackage{makecell}
\usepackage{algorithm, algpseudocode}

\newcommand{\vect}[1]{\boldsymbol{\mathbf{#1}}}
\newcommand{\shortname}{Geo4D\xspace}
\newcommand{\suppmat}{Supp. Mat\xspace}

\makeatletter
\renewcommand{\paragraph}{%
  \@startsection{paragraph}{4}%
  {\z@}{-0.25em}{-0.5em}%
  {\normalfont\normalsize\bfseries}%
}
\makeatother

\definecolor{iccvblue}{rgb}{0.21,0.49,0.74}
\definecolor{myPink}{RGB}{255,105,180}
\usepackage[pagebackref,breaklinks,colorlinks,allcolors=iccvblue,urlcolor=myPink]{hyperref}

\usepackage{amssymb}
\usepackage{pifont}
\usepackage[accsupp]{axessibility}
\newcommand{\cmark}{\ding{51}}%

\def\paperID{1776} 
\def\confName{ICCV}
\def\confYear{2025}

\title{\shortname: Leveraging Video Generators for Geometric 4D Scene Reconstruction}

\newcommand\rurl[1]{%
  \href{https://#1}{\nolinkurl{#1}}%
}

\author{Zeren Jiang$^{1}$
\quad
Chuanxia Zheng$^{1}$
\quad
Iro Laina$^{1}$
\quad
Diane Larlus$^{2}$
\quad
Andrea Vedaldi$^{1}$ \\
 $^1$Visual Geometry Group, University of Oxford \quad
 $^2$Naver Labs Europe \\
{\tt\small \{zeren, cxzheng, iro, vedaldi\}@robots.ox.ac.uk \quad diane.larlus@naverlabs.com}\\[0.1em]
\small\rurl{geo4d.github.io}
}

\begin{document}

\twocolumn[{%
\renewcommand\twocolumn[1][]{#1}%
\maketitle
\begin{center}
\captionsetup{type=figure}
\includegraphics[width=\linewidth]{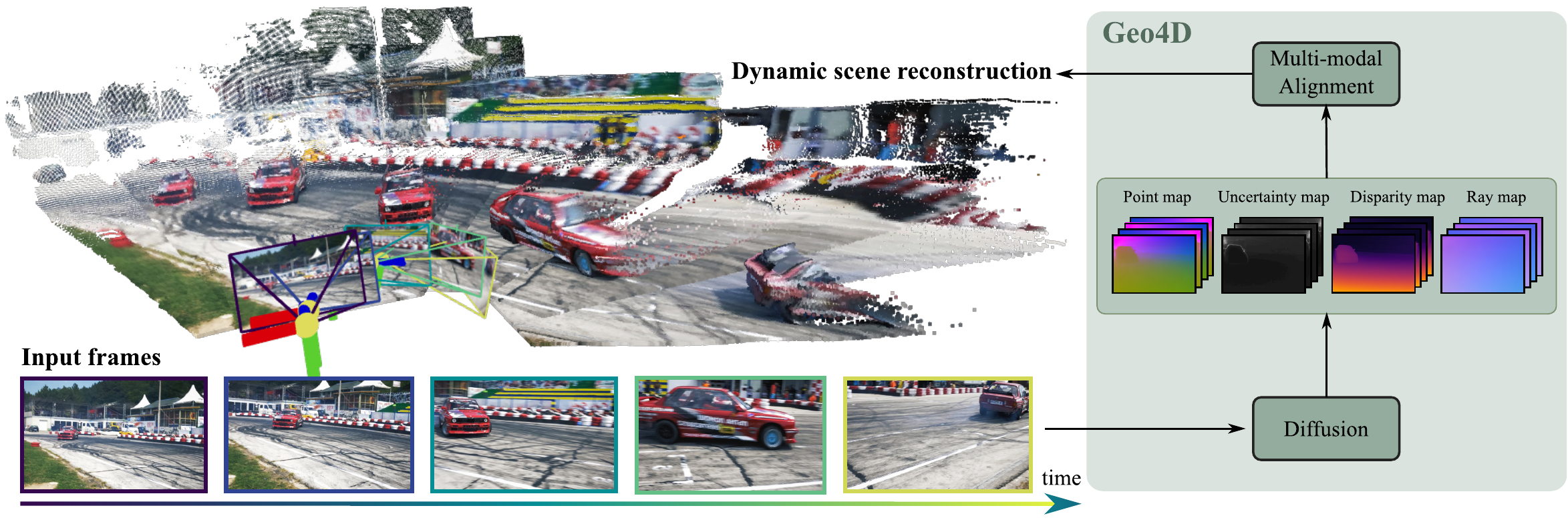}
\vspace{-0.8cm}
\captionof{figure}{
\shortname repurposes a video diffusion model~\cite{xing2024dynamicrafter} for monocular 4D reconstruction.
It uses only synthetic data for training, yet generalizes well to \emph{out-of-domain} real videos.
It predicts several geometric modalities, including point maps, disparity maps, and ray maps, fusing and aligning them to obtain state-of-the-art dynamic reconstruction even for scenes with extreme object and camera motion.
}%
\label{fig:teaser}
\end{center}%
}]

\newcommand{\tableVideoDepth}{
\begin{table*}[t]
\centering
\small
\begin{tabular}{@{}cl cc cc cc@{}}
\toprule
\multirow{2}*{\textbf{Category}} & \multirow{2}*{\textbf{Method}} & \multicolumn{2}{c}{\textbf{Sintel}~\cite{Butler2012ANO}} & \multicolumn{2}{c}{\textbf{Bonn}~\cite{Palazzolo2019ReFusion3R}} & \multicolumn{2}{c}{\textbf{KITTI}~\cite{geiger13vision}} \\ 
\cmidrule(lr){3-4}
\cmidrule(lr){5-6}
\cmidrule(lr){7-8}
  &   & Abs Rel~$\downarrow$ & $\delta < 1.25$~$\uparrow$ & Abs Rel~$\downarrow$ & $\delta < 1.25$~$\uparrow$ & Abs Rel~$\downarrow$ & $\delta < 1.25$~$\uparrow$\\ 
\midrule
\multirow{2}*{Single-frame depth} & Marigold~\cite{ke2024repurposing} & 0.532 & 51.5 & 0.091 & 93.1 & 0.149 & 79.6  \\
  & Depth-Anything-V2~\cite{lihe24DAV2} & 0.367 & 55.4 & 0.106 & 92.1 & 0.140 & 80.4  \\
\midrule
 \multirow{3}*{Video depth} & NVDS~\cite{wang23NVDS} & 0.408 & 48.3 & 0.167 & 76.6 & 0.253 & 58.8 \\
  & ChronoDepth~\cite{shao24learning} & 0.687 & 48.6 & 0.100 & 91.1 & 0.167 & 75.9 \\
  & DepthCrafter*~\cite{hu25depthcrafter} & \underline{0.270} & \underline{69.7} & 0.071 & \textbf{97.2} & \underline{0.104} & \underline{89.6} \\
\midrule
 \multirow{4}*{\makecell[c]{Video depth \\ \& Camera pose}} & Robust-CVD~\cite{Kopf2020RobustCV} & 0.703 & 47.8 &---&---&---&---\\
  & CasualSAM~\cite{Zhang2022StructureAM} & 0.387 & 54.7 & 0.169 & 73.7 & 0.246 & 62.2 \\
  & MonST3R~\cite{zhang24monst3r} & 0.335 & 58.5 & \underline{0.063} & 96.4 & \underline{0.104} & 89.5 \\
  & \textbf{Geo4D (Ours)} & \textbf{0.205} & \textbf{73.5} & \textbf{0.059} & \textbf{97.2} & \textbf{0.086} & \textbf{93.7} \\
\bottomrule
\end{tabular}
\vspace{-0.3cm}
\caption{\textbf{Video depth estimation} on Sintel~\cite{Butler2012ANO}, Bonn~\cite{Palazzolo2019ReFusion3R} and KITTI~\cite{geiger13vision} datasets.
We follow the evaluation protocols established in recent MonST3R~\cite{zhang24monst3r} for a fair comparison.
Notably, results for DepthCrafter* are reported from its latest version (v1.0.1).
The \textbf{Best} and the \underline{second best} results are highlighted.
}%
\label{tab:video_depth}
\vspace{-0.5cm}
\end{table*}
}

\newcommand{\tableAblation}{
\begin{table*}[tb!]
\centering
\small
\resizebox{\textwidth}{!}{%
\begin{tabular}{@{}ccc ccc cc ccc@{}}
\toprule
\multicolumn{3}{c}{\textbf{Training}} & \multicolumn{3}{c}{\textbf{Inference}} & \multicolumn{2}{c}{\textbf{Video Depth}} & \multicolumn{3}{c}{\textbf{Camera Pose}} \\
\cmidrule(lr){1-3}
\cmidrule(lr){4-6}
\cmidrule(lr){7-8}
\cmidrule(lr){9-11}
Point Map & Disparity Map & Ray Map & Point Map & Disparity Map & Ray Map & Abs Rel~$\downarrow$ & $\delta < 1.25$~$\uparrow$ & ATE~$\downarrow$ & RPE trans~$\downarrow$ & RPE rot~$\downarrow$\\
\midrule
\cmark &---&---& \cmark &---&---& 0.232 & 71.3 & 0.335 & 0.076 & 0.731    \\
\cmark & \cmark & \cmark & \cmark &---&--- & 0.223 & 72.5 & 0.237 & 0.070 & 0.566   \\
\cmark & \cmark & \cmark &---& \cmark &--- & 0.211 & 73.4 &---&---&---  \\
\cmark & \cmark & \cmark &---&---& \cmark  &---&---& 0.268 & 0.192 & 1.476   \\
\cmark & \cmark & \cmark & \cmark & \cmark & \cmark & \textbf{0.205} & \textbf{73.5} & \textbf{0.185} & \textbf{0.063} & \textbf{0.547}    \\
\bottomrule
\end{tabular}
}
\vspace{-0.3cm}
\caption{\textbf{Ablation study for the different modalities of the geometric representation} on the Sintel~\cite{Butler2012ANO} dataset.
We demonstrate the effectiveness of our key design choices that both leverage multi-modality as additional training supervision signal and postprocess through our proposed multi-modal alignment algorithm will improve the overall performance.}%
\label{tab:ablation}
\vspace{-0.5cm}
\end{table*}
}

\newcommand{\tableCamerapose}{
\begin{table}[tb!]
\centering
\small
\tabcolsep=0.4em
\resizebox{\linewidth}{!}{%
\begin{tabular}{@{}l ccc ccc@{}}
\toprule
\multirow{2}*{\textbf{Method}} & \multicolumn{3}{c}{\textbf{Sintel}} & \multicolumn{3}{c}{\textbf{TUM-dynamics}} \\ 
\cmidrule(lr){2-4}\cmidrule(lr){5-7}
  & ATE~$\downarrow$& RPE-T~$\downarrow$& RPE-R~$\downarrow$& ATE~$\downarrow$& RPE-T~$\downarrow$& RPE-R~$\downarrow$\\ 
\midrule
Robust-CVD~\cite{Kopf2020RobustCV} & 0.360 & 0.154 & 3.443 & 0.153 & 0.026 & 3.528  \\
CasualSAM~\cite{Zhang2022StructureAM} & 0.141 & \textbf{0.035} & 0.615 & 0.071 & 0.010 & 1.712  \\
MonST3R~\cite{zhang24monst3r} & \textbf{0.108} & 0.042 & 0.732 & \textbf{0.063} & \textbf{0.009} & 1.217  \\
\textbf{Geo4D (Ours)} & 0.185 & 0.063 & \textbf{0.547} & 0.073 & 0.020 & \textbf{0.635} \\
\bottomrule
\end{tabular}
}
\vspace{-0.3cm}
\caption{\textbf{Quantitative evaluation for camera pose estimation.} We achieve comparable camera pose estimation performance with other discriminative SOTA methods.}%
\label{tab:camera}
\vspace{-0.5cm}
\end{table}
}

\newcommand{\tableStride}{
\begin{table}[tb!]
\centering
\small
\resizebox{\linewidth}{!}{%
\begin{tabular}{@{}c c cc ccc@{}}
\toprule
\multirow{2}*{\textbf{Stride}} & \multirow{2}*{\textbf{s / frame}} & \multicolumn{2}{c}{\textbf{Video Depth}} & \multicolumn{3}{c}{\textbf{Camera Pose}} \\ 
\cmidrule(lr){3-4}
\cmidrule(lr){5-7}
  &   & Abs Rel~$\downarrow$ & $\delta < 1.25$~$\uparrow$ & ATE~$\downarrow$ & RPE trans~$\downarrow$ & RPE rot~$\downarrow$\\ 
\midrule

15 & \textbf{0.92} & 0.213 & 72.4 & 0.210 & 0.092 & 0.574    \\
8 & 1.24 & 0.212 & 72.8 & 0.222 & 0.074 & 0.524   \\
4 & 1.89 & 0.205 & \textbf{73.5} & 0.185 & 0.063 & 0.547   \\
2 & 3.26 & \textbf{0.204} & 72.9 & \textbf{0.181} & \textbf{0.058} & \textbf{0.518}   \\
\bottomrule
\end{tabular}
}
\vspace{-0.3cm}
\caption{\textbf{Ablation study for the  temporal sliding window stride} on the Sintel~\cite{Butler2012ANO} dataset. There is a trade-off between performance and inference speed.}%
\label{tab:ablation_S}
\vspace{-0.5cm}
\end{table}
}
\newcommand{\tableDataset}{
\begin{table}[tb!]
\centering
\small
\resizebox{\linewidth}{!}{%
\begin{tabular}{@{}ccccc@{}}
\toprule
Dataset & Scene type & \#Frames & \#Sequences & Ratio \\
\midrule
PointOdyssey~\cite{zheng2023pointodyssey} & Indoors/Outdoors & 200K & 131 & 16.7\%  \\
TartanAir~\cite{Wang2020TartanAirAD} & Indoors/Outdoors & 1000K & 163 & 16.7\%  \\
Spring~\cite{Mehl2023_Spring} & Outdoors & 6K & 37 & 16.7\%  \\
VirtualKITTI~\cite{cabon2020vkitti2} & Driving & 43K & 320 & 16.7\%  \\
BEDLAM~\cite{black2023bedlamsyntheticdatasetbodies} & Indoors/Outdoors & 380K & 10K & 33.3\%  \\
\bottomrule
\end{tabular}
}
\vspace{-0.3cm}
\caption{\textbf{Details of training datasets.} Our method only uses synthetic datasets for training.}%
\label{tab:ablation_Dataset}
\end{table}
}

\newcommand{\tableVAE}{
\begin{table}[tb!]
\centering
\small
\resizebox{\linewidth}{!}{%
\begin{tabular}{@{}c cc ccc@{}}
\toprule
\multirow{2}*{\textbf{Method}} & \multicolumn{2}{c}{\textbf{Video Depth}} & \multicolumn{3}{c}{\textbf{Camera Pose}} \\ 
\cmidrule(lr){2-3}
\cmidrule(lr){4-6}
   & Abs Rel~$\downarrow$ & $\delta < 1.25$~$\uparrow$ & ATE~$\downarrow$ & RPE trans~$\downarrow$ & RPE rot~$\downarrow$\\ 
\midrule
w/o fine-tuned & 0.212 & 72.1 & 0.192 & \textbf{0.061} & 0.577   \\
w fine-tuned & \textbf{0.205} & \textbf{73.5} & \textbf{0.185} & 0.063 & \textbf{0.547}   \\
\bottomrule
\end{tabular}
}
\vspace{-0.3cm}
\caption{\textbf{Ablation study for the fine-tuned point map VAE} on the Sintel~\cite{Butler2012ANO} dataset.
The fine-tuned point map VAE performs better than the original one.}%
\label{tab:ablation_VAE}
\end{table}
}

\newcommand{\tableStep}{
\begin{table}[tb!]
\centering
\small
\resizebox{\linewidth}{!}{%
\begin{tabular}{@{}c cc ccc@{}}
\toprule
\multirow{2}*{\textbf{Steps}} & \multicolumn{2}{c}{\textbf{Video Depth}} & \multicolumn{3}{c}{\textbf{Camera Pose}} \\ 
\cmidrule(lr){2-3}
\cmidrule(lr){4-6}
& Abs Rel~$\downarrow$ & $\delta < 1.25$~$\uparrow$ & ATE~$\downarrow$ & RPE trans~$\downarrow$ & RPE rot~$\downarrow$\\ 
\midrule
1 & 0.221 & 70.7 & 0.234 & 0.072 & 0.753    \\
5 & \textbf{0.205} & \textbf{73.5} & \textbf{0.185} & \textbf{0.063} & 0.547    \\
10 & 0.207 & 73.2 & 0.212 & 0.071 & \textbf{0.508}   \\
25 & 0.220 & 72.2 & 0.211 & 0.074 & 0.564   \\
\bottomrule
\end{tabular}
}
\vspace{-0.3cm}
\caption{\textbf{Ablation study for the DDIM sampling steps.} on the Sintel~\cite{Butler2012ANO} dataset. }%
\label{tab:ablation_step}
\end{table}
}

\newcommand{\tableFF}{
\begin{table}[t]
\centering
\small
\renewcommand*{\arraystretch}{0.85}
\resizebox{\linewidth}{!}{%
\begin{tabular}{@{}c cc ccc@{}}
\toprule
\multirow{2}*{\textbf{Method}} & \multicolumn{2}{c|}{\textbf{Video Depth}} & \multicolumn{3}{c}{\textbf{Camera Pose}} \\
\cmidrule(lr){2-3}
   & Abs Rel~$\downarrow$ & $\delta < 1.25$~$\uparrow$ & ATE~$\downarrow$ & RPE trans~$\downarrow$ & RPE rot~$\downarrow$\\ 
\midrule
MonST3R & 0.324 & 55.6 & 0.163 & 0.099 & 0.632   \\
Ours & \textbf{0.206} & \textbf{73.1} & \textbf{0.086} & \textbf{0.086} & \textbf{0.443}   \\
\bottomrule
\end{tabular}
}
\caption{\textbf{Feed-forward evaluation} on Sintel.}%
\label{tab:rebuttal_FF}
\end{table}
}

\newcommand{\tablePM}{
\begin{table}[t]
\centering
\small
\resizebox{\linewidth}{!}{%
\begin{tabular}{@{}l l cc ccc@{}}
\toprule
\multirow{2}*{\textbf{Training}} & \multirow{2}*{\textbf{Inference}} & \multicolumn{2}{c|}{\textbf{Video Depth}} & \multicolumn{3}{c}{\textbf{Camera Pose}} \\
\cmidrule(lr){2-3}
  &    & Abs Rel~$\downarrow$ & $\delta < 1.25$~$\uparrow$ & ATE~$\downarrow$ & RPE-T~$\downarrow$ & RPE-R~$\downarrow$\\ 
\midrule
DM+RM & DM+RM & 0.211 & 71.2 & 0.322 & 0.215 & 2.007   \\
DM+RM+PM & DM+RM & 0.211 & 73.4 & 0.268 & 0.192 & 1.476   \\
DM+RM+PM & DM+RM+PM & \textbf{0.205} & \textbf{73.5} & \textbf{0.185} & \textbf{0.063} & \textbf{0.547}   \\
\bottomrule
\end{tabular}
}
\vspace{-0.3cm}
\caption{\textbf{Effectiveness of point maps} on Sintel.}%
\label{tab:rebuttal_PM}
\vspace{-0.8cm}
\end{table}
}

\newcommand{\algopt}{
\begin{algorithm}[tb!]
\small
\caption{Multi-Modal Alignment Optimization}%
\label{alg:A}
\begin{algorithmic}[1]
\State{$\bm{X}^{i,g},
\bm{D}^{i,g},
\bm{r}^{i,g}
\gets
$
Predicted by our diffusion model}

\State{$\bm{D}^i_\text{p},
\lambda^g_\text{p},
\bm{R}^g_\text{p},
\bm{\beta}^g_\text{p}
\gets
$ 
Initialized by Umeyama algorithm}

\State{
$\textbf{K}^k_\text{p} \gets$
Optimized from $\vect{X}^{k,g^k}$}
\State{
$\mathbf{R}^{i}_\text{p},
\mathbf{o}^{i}_\text{p}\gets
$ 
Initialized by Ransac PnP from pointmaps
$\bm{X}^i$}

\State{
$\mathbf{R}^{i,g}_\text{c},
\bm{o}^{i,g}_\text{c}\gets$
Initialized by~\cref{eq:opt_cam1,eq:opt_cam2} from raymaps
$\bm{r}^{i,g}$}

\Repeat
\If{Iteration = Align start iteration}

\State{
$\lambda^g_\text{d}, \beta^g_\text{d}
\gets
\arg\min
\mathcal{L}_{\text{d}}
$ (~\cref{eq:align_depth})}

\State{
$\bm{R}^g_\text{c},
\lambda^g_\text{c},
\bm{\beta}^g_\text{c}
\gets
\arg\min
\mathcal{L}_{\text{c}}
$ (~\cref{eq:align_cam})}

\ElsIf{Iteration $<$ Align start iteration}

    \State{
    $\bm{D}^i_\text{p},
    \mathbf{K}^i_\text{p},
    \mathbf{R}^i_\text{p},
    \bm{o}^i_\text{p},
    \lambda^g_\text{p},
    \bm{R}^g_\text{p},
    \bm{\beta}^g_\text{p},
    \gets
    \arg\min
    \mathcal{L}_{\text{p}}
    + \mathcal{L}_{\text{s}} $ }
    
\Else

    \State{
    $\bm{D}^i_\text{p},
    \mathbf{K}^i_\text{p},
    \mathbf{R}^i_\text{p},
    \bm{o}^i_\text{p},
    \lambda^g_{*},
    \bm{R}^g_{*},
    \bm{\beta}^g_{*}
    \gets
    \arg\min
    \mathcal{L}_{\text {all } }$}
    
\EndIf
\Until{max loop reached}
\end{algorithmic}
\end{algorithm}
}

\begin{abstract}
We introduce \shortname, a method to repurpose video diffusion models for monocular 3D reconstruction of dynamic scenes.
By leveraging the strong dynamic priors captured by large-scale pre-trained video models, \shortname can be trained using only synthetic data while generalizing well to real data in a zero-shot manner.
\shortname predicts several complementary geometric modalities, namely point, disparity, and ray maps.
We propose a new multi-modal alignment algorithm to align and fuse these modalities, as well as a sliding window approach at inference time, thus enabling robust and accurate 4D reconstruction of long videos.
Extensive experiments across multiple benchmarks show that \shortname significantly surpasses state-of-the-art video depth estimation methods.
\end{abstract}

\section{Introduction}%
\label{sec:intro}

We consider the problem of \emph{feed-forward 4D reconstruction}, which involves learning a neural network to reconstruct the 3D geometry of a dynamic scene from a monocular video.
This task is particularly challenging for videos captured in uncontrolled settings, such as those shot with handheld cameras or downloaded from the Internet.
However, a robust solution to this problem would have a tremendous impact on a wide range of applications, from video understanding to computer graphics and robotics.

4D reconstruction from videos is related to multi-view static 3D reconstruction, which is typically addressed using methods from visual geometry like bundle adjustment.
Recent neural networks~\cite{wang2024dust3r,jianyuan-wang25vggt:} have emerged as powerful tools that can replace, or at least complement, bundle adjustment.
They excel especially in difficult reconstruction scenarios, involving, \eg, textureless surfaces and occlusions, thanks to the priors they learn from data.
Given the additional challenges involved in 4D reconstruction, we expect that such priors would benefit this task even more.

In fact, powerful networks like DUSt3R~\cite{wang2024dust3r}, designed for static multi-view 3D reconstruction, have recently been extended to the dynamic case, for example by MonST3R~\cite{zhang24monst3r}.
However, these models are heavily engineered to solve specific 3D reconstruction problems.
Most importantly, they require significant amounts of training data with 3D annotations for supervision.
Such data is difficult to collect for dynamic scenes, especially in real life.
This suggests using 4D synthetic training data instead.
However, this data is difficult to obtain at scale, and the gap with the real world can compromise generalization.

One way to mitigate this problem is to pre-train the model on tasks related to 3D reconstruction for which real data is easily available.
For example, DUSt3R~\cite{wang2024dust3r} and derived methods~\cite{zhang24monst3r} use image matching for pre-training~\cite{weinzaepfel22croco:}.
Here, we suggest starting instead from an off-the-shelf \emph{video generator}.
Video generators are powerful models, often considered proxies of world simulators~\cite{li24sora,nvidia25cosmos,parker-holder24genie}.
More importantly for us, the videos they generate demonstrate an understanding of effects like camera motion and perspective, as well as typical object motion in the context of a scene.
However, they only generate pixels, leaving any 3D or 4D understanding \emph{implicit} and thus not directly actionable.

In this work, we show that \textit{a pre-trained off-the-shelf video generator} can be turned into an effective \textit{monocular feed-forward 4D reconstructor}.
To this end, we introduce \textbf{\shortname}, a novel approach for adapting Video Generators for \textbf{Geo}metric \textbf{4D} Reconstruction.
With \shortname, we demonstrate that these generic video architectures can successfully solve complex 4D reconstruction tasks, which is a step towards future video foundation models that natively integrate 4D geometry.
Prior work such as Marigold~\cite{ke2024repurposing} and concurrent work DepthCrafter~\cite{hu25depthcrafter} have looked at adapting, respectively, image and video generators for depth estimation.
Here, we go one step further and consider \textit{the full recovery of 4D geometry, including camera motion and dynamic 3D structure}.

With \shortname, our goal is to make 4D geometry explicit in the video generator.
This in turn requires us to choose an \emph{explicit representation} of 4D information.
We follow DUSt3R and adopt its viewpoint-invariant point maps.
Namely, we associate each pixel in each frame with the coordinate of the corresponding 3D point, expressed relative to the first frame in the video, used as a reference.
Hence, the static parts of the point clouds extracted from the different frames line up, and the dynamic parts form a 3D `trace' of the motion of the dynamic objects, as shown in~\cref{fig:teaser}.

Viewpoint-invariant point maps are a powerful representation because they implicitly encode the camera motion and intrinsics and can be easily predicted by a neural network~\cite{wang2024dust3r}.
However, they are not necessarily the best representation for all parts of the scene, particularly for points far away from the observer or even at infinity, such as the sky.
We thus consider two more \emph{modalities} with better dynamic range, namely disparity maps and camera ray maps.
Ray maps, in particular, are defined for all image pixels regardless of the scene geometry.

Our model thus predicts three modalities:
point, disparity, and ray maps.
These modalities are redundant in principle, but complementary in practice.
At test time, we reconcile them via a fast, global optimization step and show that this leads to significantly more robust 4D reconstructions.
Due to depth and ray map prediction, we show very strong empirical results on video depth estimation and in the recovery of the camera orientation.

One of the challenges of monocular 4D reconstruction is that it is ambiguous, significantly more so than static 3D reconstruction.
However, the stochastic nature of the video generator can help deal with this ambiguity.
We also introduce uncertainty maps in the encoder-decoder architecture that processes the geometric maps, and integrate them into the multi-modal alignment process.

Overall, our contributions are as follows.
(i) We introduce \shortname, a 4D feed-forward network for dynamic scene reconstruction that builds on top of an off-the-shelf video generator.
(ii) We suggest generating multiple partially redundant geometric modalities and fusing them at test time via lightweight optimization.
(iii) We show the benefits of this multi-modal fusion in terms of improved 4D prediction accuracy.
Experiments show that this model can reconstruct even highly dynamic scenes (such as the drifting scene in DAVIS~\cite{Huang_2016} presented in \cref{fig:teaser}) and outperforms current video depth and camera rotation estimation methods.

\section{Related Work}%
\label{sec:rel}

\subsection{Dynamic Scene Reconstruction}

\paragraph{Static 3D reconstruction.}

Feed-forward 3D reconstruction has achieved remarkable success across various representations, including voxels~\cite{choy20163d,tulsiani2017multi,sitzmann2019deepvoxels}, meshes~\cite{wang2018pixel2mesh,gkioxari2019mesh,siddiqui2024meshgpt}, and point clouds~\cite{lin2018learning,yu2021pointr}.
These advancements have been further driven by implicit neural representations~\cite{park2019deepsdf,peng2020convolutional,mildenhall2020nerf,sitzmann2021light} and the emergence of 3D Gaussian Splatting (3D-GS)~\cite{kerbl20233d,charatan2024pixelsplat,szymanowicz24splatter,chen24mvsplat:,szymanowicz25flash3d:,smart2024splatt3r}.
Recently, DUSt3R~\cite{wang2024dust3r} introduced a point map representation for scene-level 3D reconstruction, followed by~\cite{wang20243d,leroy2024grounding,Yang_2025_Fast3R,jianyuan-wang25vggt:}.
However, these models predominantly focus on \emph{static} 3D reconstruction.
Our approach also uses point maps as a representation but extends them to handle \emph{dynamic} scenes, which present additional challenges due to object motion over time.

\begin{figure*}[t]
    \centering
    \includegraphics[width=\textwidth]{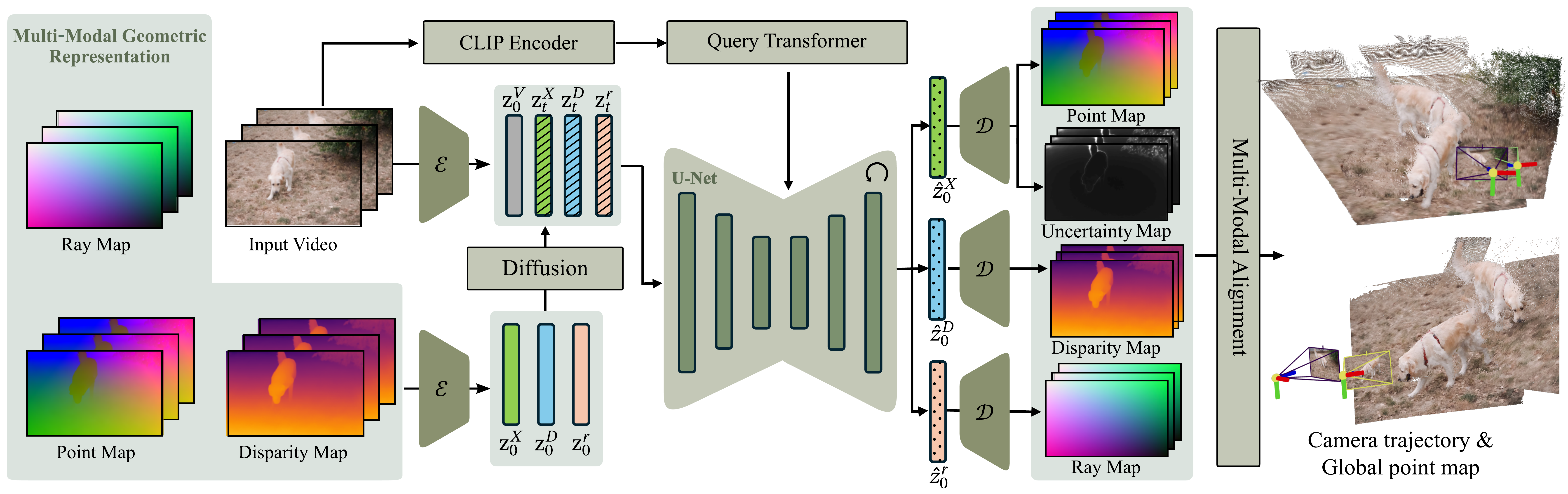}
    \vspace{-0.8cm}
    \caption{\textbf{Overview of \shortname.} During training, 
    video conditions are injected by locally concatenating the latent feature of the video with diffused geometric features $\vect{z}_t^{\vect{X}}, \vect{z}_t^{\vect{D}}, \vect{z}_t^{\vect{r}}$ and are injected globally via cross-attention in the denoising U-Net, after CLIP encoding and a query transformer. The U-Net is fine-tuned via Eq.~\ref{eq:vdm}. During inference, iteratively denoised latent features $\hat{\vect{z}}_0^{\vect{X}}, \hat{\vect{z}}_0^{\vect{D}}, \hat{\vect{z}}_0^{\vect{r}}$ are decoded by the fine-tuned VAE decoder, followed by multi-modal alignment optimization for coherent 4D reconstruction.} \label{fig:framework}
\vspace{-0.3cm}
\end{figure*}

\paragraph{Iterative 4D reconstruction.}

Iterative or optimization-based approaches reconstruct 4D models from monocular videos by iteratively fitting the observed data.
Classical techniques often rely on RGB-D sensors~\cite{newcombe2015dynamicfusion,innmann2016volumedeform}, but such steps are impractical for many real-world scenes.
Recently, with advancements in neural representations~\cite{park2019deepsdf,mildenhall2020nerf}, NeRF-based approaches~\cite{park2021nerfies,park2021hypernerf,li2021neural,pumarola2021d,li2023dynibar,jiang2024multiply} have shown impressive results.
However, volume rendering in NeRF is computationally expensive.
Convergence and rendering speed can be improved by using 3D-GS representations~\cite{kerbl20233d,wu24_4DGS,yang2024deformable,som24wang,dreamscene4d,lin24gaussian,lei2024mosca,yuan20251000fps4dgaussian}, which reduce but do not eliminate the cost of iterative optimization.
Very recently, MegaSaM~\cite{li2024_megasam} achieved highly accurate and robust camera pose estimation and reconstruction for dynamic videos, but it requires accurate monocular depth priors.
Similarly, Uni4D~\cite{yao2025uni4dunifyingvisualfoundation} produces accurate 4D reconstructions by leveraging various visual foundation models and performing multi-stage bundle adjustment.
In contrast, our approach is a diffusion-driven feed-forward framework, which eliminates the need for per-video bundle adjustment and external depth estimation models.

\paragraph{Feed-forward 4D reconstruction.}

Similar to our approach, recent works have started to explore \emph{feed-forward} 4D reconstruction for dynamic scenes: a monocular video with dynamic objects is processed by a neural network to recover a 4D representation.
For objects, L4GM~\cite{ren2024l4gm} and Animate3D~\cite{jiang2024animated} first generate multi-view videos from a monocular video input, and subsequently apply 3D-GS~\cite{kerbl20233d} to reconstruct a temporally consistent 4D model.
For scenes, a notable example is MonST3R~\cite{zhang24monst3r}, which adapts the \emph{static} scene reconstruction of DUSt3R~\cite{wang2024dust3r} to handle \emph{dynamic} scenes.
Very recently, Easi3R~\cite{chen2025easi3r} applies attention adaptation during inference and performs 4D reconstruction based on DUSt3R~\cite{wang2024dust3r} in an efficient, training-free manner.

\subsection{Geometric Diffusion Models}

Our method builds upon advancements in video diffusion models~\cite{ho2022video,wang2023modelscope,singer2023makeavideo,blattmann2023align,ge2023preserve,wang2023videocomposer,guo2024animatediff,blattmann2023stable,zhang2024show,xing2024dynamicrafter,kong2024hunyuanvideo}, which generate \emph{temporally consistent} videos from text or image prompts.
Recent studies have explored the rich 3D priors embedded within large-scale pre-trained diffusion models, employing either knowledge distillation~\cite{pooledreamfusion,wang2023score,lin2023magic3d,wang2024prolificdreamer,jakab2024farm3d,melas2023realfusion} or fine-tuning~\cite{liu2023zero,shi2024mvdream,li2024instant3d,liu2024syncdreamer,han2024vfusion3d,zheng2024free3d,long2024wonder3d,voleti2025sv3d} for 3D reconstruction and generation.
While these methods have significantly advanced single-object 3D reconstruction from sparse inputs, they remain largely constrained to \emph{static, isolated} objects centered within an image.
Beyond single-object reconstruction, several recent efforts have extended pre-trained diffusion models to tackle scene-level 3D tasks, such as optical flow estimation~\cite{saxena2023surprising}, view synthesis~\cite{gao2024cat3d,sargent2024zeronvs,chen2024mvsplat360,yu2024viewcrafter,liu2024reconxreconstructscenesparse,szymanowicz2025bolt3d}, depth estimation~\cite{duan2024diffusiondepth,ke2024repurposing,zhao2023unleashing}, and normal estimation~\cite{qiu2024richdreamer,fu2024geowizard,krishnan2025orchid}.
More related to our approach, Matrix3D~\cite{lu2025matrix3d} jointly predicts depth and camera parameters, and WVD~\cite{zhang2024world} introduces a hybrid RGB+point map representation for scene reconstruction.
However, these approaches assume \emph{static} 3D environments, whereas we address \emph{dynamic} 4D scene reconstruction, which is a much harder problem due to object motion across time.

More closely related to our approach, concurrent GeometryCrafter~\cite{xu2025geometrycrafterconsistentgeometryestimation} introduced a point map VAE with a dual encoder-decoder architecture to improve reconstruction accuracy.
However, their point maps are defined in individual camera coordinates, necessitating the use of additional segmentation~\cite{kirillov2023segment} and tracking models~\cite{SpatialTracker} to recover the global point map and estimate camera poses.
Aether~\cite{aether}, on the other hand, outputs depth maps and ray maps from a video diffusion model for 4D reconstruction.
In contrast, our experiments demonstrate that performance can be significantly enhanced by jointly predicting multiple geometric modalities that capture diverse dynamic ranges, ensuring better temporal coherence and robustness.
Importantly, our approach is self-contained and does \emph{not} rely on external models, enhancing its generality and reliability.

\section{Method}%
\label{sec:method}

Our goal is to learn a neural network $f_{\bm \theta}$ that can reconstruct dynamic 3D scenes from monocular videos.
Given as input a monocular video
$\mathcal{I}=\{\bm{I}^i\}_{i=1}^N$
consisting of $N$ frames, where each frame is an RGB image $\bm{I}^i\in\mathbb{R}^{H\times W\times3}$,
the network $f_{\bm \theta}$ returns a representation of its 4D geometry:
\begin{equation}
\label{eq:goal}
f_{\bm \theta}:
\{\bm{I}^i\}_{i=1}^N
\mapsto
\{(\bm{D}^i,\bm{X}^i,\bm{r}^i)\}_{i=1}^N.
\end{equation}
The network computes the disparity map
$\bm{D}^i\in\mathbb{R}^{H\times W\times1}$,
the viewpoint-invariant point map
$\bm{X}^i\in\mathbb{R}^{H\times W\times3}$,
and the ray map
$\bm{r}^i\in\mathbb{R}^{H\times W\times6}$
for each frame $\bm{I}^i$, $i = 1,\dots,N$.
As we discuss in \cref{sec:method_4ddiffusion}, these quantities collectively represent the 4D geometry of a scene, including its dynamic structure and time-varying camera extrinsic and intrinsic parameters.
No camera parameters are provided as input; these are implicitly estimated by the model as well.

We implement $f_{\bm{\theta}}$ as a video diffusion model, where ${\bm{\theta}}$ are the learnable parameters.
We discuss the relevant background on video diffusion models in \cref{sec:pre}.
Then, in \cref{sec:method_4ddiffusion}, we describe how we extend the model to predict the three modalities of the 4D geometry.
Finally, in \cref{sec:mml}, we describe how we fuse and align these modalities to obtain a coherent 4D reconstruction at test time.

\subsection{Preliminaries: Video Diffusion Model}%
\label{sec:pre}

Our key insight is that by building on pre-trained video diffusion models, our approach can exploit the strong motion and scene geometry priors inherently encoded within these models.
Specifically, we build \shortname on top of DynamiCrafter~\cite{xing2024dynamicrafter}, a ``foundation'' video diffusion model.
DynamiCrafter is a latent diffusion model~\cite{rombach2022high}:
it uses a variational autoencoder (VAE) to obtain a more compact video representation and thus reduce computational complexity.
During training, a target sequence
$\mathcal{X}=\bm{x}^{1:N}$
is first encoded into the latent space using the encoder
$\bm{z}_0^{1:N}=\mathcal{E}(\bm{x}^{1:N})$,
and then perturbed by
$\bm{z}_t^{1:N}
=
\sqrt{\bar{\alpha_t}}\bm{z}_0^{1:N}+
\sqrt{1-\bar{\alpha_t}}\epsilon^{1:N}
$,
where $\epsilon\sim\mathcal{N}(\bm{0},\bm{I})$ is Gaussian noise, and $\bar{\alpha_t}$ is the noise level at step $t$ of $T$ noising steps.
The denoising network
$\epsilon_{\bm\theta}$
is then trained to reverse this noising process by optimizing the following objective: 
\begin{equation}
\label{eq:vdm}
\min_{\bm\theta}
\mathbb{E}_{(\bm{x}^{1:N},y), t,\epsilon^{1:N} \sim \mathcal{N}(\bm{0}, \bm{I})}
\left \|\epsilon^{1:N}-\epsilon_{\bm\theta}\left(\bm{z}_t^{1:N}, t,y\right)\right \|_2^2,
\end{equation}
where $y$ is the conditional input.
Once trained,
the model generates a video prompted by
$y$
via iteratively denoising from pure noise
$\bm{z}_T^{1:N}$,
and then decoding the denoised latent with a decoder
$\mathcal{\hat{X}}=\mathcal{D}(\hat{\bm{z}}_0^{1:N})$.

\subsection{Multi-modal Geometric 4D Diffusion}%
\label{sec:method_4ddiffusion}

We first provide a more precise description of the 4D multi-modal representation output by our model, and then explain how it is encoded in the latent space for generation.

\paragraph{Multi-modal geometric representations.}

The dynamic 3D structure of a scene is represented by a sequence of point maps
$\{\bm{X}^i\}_{i=1}^N$,
one for each of its $N$ frames.
Let $(u,v)$ denote the pixel coordinates in the image plane.
Then, the value $X^i_{uv} \in \mathbb{R}^3$ is the 3D coordinate of the scene point that lands at pixel $(u,v)$ in frame $\bm{I}^i$, expressed in the reference frame of camera $i=1$.
Because the reference frame is fixed and independent of the time-varying viewpoint, we call these point maps \emph{viewpoint-invariant}.
The advantages of this representation are convincingly demonstrated by DUSt3R~\cite{wang2024dust3r}.
For a static scene, or by knowing which image pixels correspond to the static part of a scene, knowledge of the point maps allows recovery of the intrinsic and extrinsic camera parameters as well as the scene depth.
This is done by solving an optimization problem that aligns the dynamic point maps with a pinhole camera model.

As noted in \cref{sec:intro}, while point maps
$\{\bm{X}^i\}_{i=1}^N$ fully
encode the 4D geometry of the scene, 
they are not effective for all parts of the scene.
Their dynamic range is limited, and they are not even defined for points at infinity (\eg sky).
Hence, we consider two additional modalities: disparity maps $\{\bm{D}^i\}_{i=1}^N$ and camera ray maps $\{\bm{r}^i\}_{i=1}^N$, also encouraged by prior evidence~\cite{fu2024geowizard,krishnan2025orchid,lu2025matrix3d} that diffusion models can benefit from learning to predict multiple quantities.
Disparity maps are not viewpoint-invariant, but have a better dynamic range than point maps (the disparity is zero for points at infinity).
Ray maps represent only the camera parameters and are defined for all image pixels, independent of the scene geometry.
For the disparity map, $D^i_{uv}$ is the disparity (inverse depth) of the scene point that lands at pixel $(u,v)$, as seen in frame $\bm{I}^i$.
For the ray map, we adopt Pl{\"u}cker coordinates~\cite{sitzmann2021light,watsonnovel,zheng2024free3d}, \ie,
$\bm{r}_{uv}=(\bm{d}_{uv},\bm{m}_{uv})$,
where
$
\bm{d}_{uv}
=
\mathbf{R}^{\top}\mathbf{K}^{-1}(u, v, 1)^{\top}
$
is the ray direction,
and
$\bm{m}_{uv}
=
-\mathbf{R}^{\top}\mathbf{t}
\times
\bm{d}_{uv}$,
where
$
(\mathbf{R}, \mathbf{K}, \mathbf{t})
$
are the camera's rotation, calibration, and translation parameters.

\paragraph{Multi-modal latent encoding.}

The three modalities come in the form of images and can thus be naturally predicted by the video diffusion architecture.
However, this requires first mapping them to the latent space, for which we need suitable versions of the encoder $\mathcal{E}$ and decoder $\mathcal{D}$ from \cref{sec:pre}.
Related prior work~\cite{ke2024repurposing,fu2024geowizard} for depth prediction simply repurposes a pre-trained image encoder-decoder without modification.
We found this to work well for 
disparity and ray maps, but not for point maps.
Hence, for the point maps only, we fine-tune the pre-trained decoder $\mathcal{D}$ using the following objective function~\cite{wu2020unsupervised}:
\begin{equation}
    \label{eq:loss_vae_point}
    \mathcal{L} =
    - \sum_{uv} \ln\frac{1}{\sqrt{2}\bm{\sigma}_{uv}}
    \exp
    -\frac{\sqrt{2}\ell_1(
        \mathcal{D}(\mathcal{E}(\bm{X}))_{uv},
        \bm{X}_{uv})}{\bm{\sigma}_{uv}},
\end{equation}
where $\bm{\sigma} \in \mathbb{R}^{H \times W}$ is the uncertainty of the reconstructed point map, which is also predicted by an additional branch of our VAE decoder.
We leave the encoder $\mathcal{E}$ unchanged to modify the latent space as little as possible; instead, we normalize the point maps to the range $[-1, 1]$ to make them more compatible with the pre-trained image encoder.

\paragraph{Video conditioning.}

The original video diffusion model is conditioned on a single image, but here we need to condition it on the entire input video
$\mathcal{I}=\{\bm{I}^i\}_{i=1}^N$.
To this end, we use a hybrid conditioning mechanism with two streams.

As shown in Fig.~\ref{fig:framework}, in one stream, we extract a global representation of each frame $\bm{I}^{i}$ by passing it to CLIP~\cite{radford2021learning} followed by a lightweight learnable query transformer~\cite{awadalla2023openflamingo}.
These vectors are incorporated in the transformer via cross-attention layers injected in each U-Net block.
In the other stream, we extract \emph{local spatial} features from the VAE encoder and concatenate them channel-wise to the noised latents, encoding the generated 4D modalities $\{(\bm{D}^i,\bm{X}^i,\bm{r}^i)\}_{i=1}^N$.

\subsection{Multi-Modal Alignment}%
\label{sec:mml}

As noted, \shortname predicts several non-independent geometric modalities.
Furthermore, processing all frames of a long monocular video simultaneously with a video diffusion model is computationally prohibitive.
Therefore, during inference, we use a \emph{temporal sliding window} that segments the video into multiple overlapping clips, with partial overlap to facilitate joining them.
The goal of this section is to fuse the resulting multi-modal and multi-window data into a single, coherent reconstruction of the entire video.

\paragraph{Temporal sliding window.}

Given a video $\mathcal{I}=\{\bm{I}^i\}_{i=1}^N$ with $N$ frames, we divide it into several video clips
$
\mathcal{G} = \{g^k\}
$,
$
k\in\mathcal{S}
$,
where each clip $g^k$ contains $V$ frames $\{\bm{I}^i\}_{i=k}^{k+V-1}$,
and the set of starting indices is
$\mathcal{S}=\{0,s,2s,\dots,
\left \lfloor\frac{N-V}{s} \right \rfloor
s\}
\cup
\{N-V\}
$.
Here, $s$ is the sliding window stride.
The final term $\{N-V\}$ ensures that the last clip always includes the final frames of the video.

\paragraph{Alignment objectives.}

First, given the predicted point maps
$\bm{X}^{i,g}$
for each frame $i$ in each video clip $g \in \mathcal{G}$, we derive corresponding \emph{globally aligned} point maps in world coordinates, as well as the relative camera motion and scale parameters.
We denote these quantities with the $\mathrm{p}$ subscript to emphasize that they are inferred from the point map predictions.
To do so, we extend the pairwise global alignment loss from DUSt3R to a \emph{group-wise} one:
\begin{equation}
\label{eq:align_point}
\mathcal{L}_{\text{p}}
\left(\bm{X}, \lambda_\text{p}^g, \bm{P}_\text{p}^g\right)
=
\sum_{g \in \mathcal{G}}
\sum_{i \in g}
\sum_{uv}
\left \|
\frac{
\bm{X}^i_{uv} - \lambda_\text{p}^g
\bm{P}_\text{p}^{g}
\bm{X}^{i,g}_{uv}
}{{\bm\sigma}_{uv}^{i,g}}
\right \|_1,
\end{equation}
where $\lambda_\text{p}^g$
and 
$
\bm{P}_\text{p}^{g}
=
[\mathbf{R}_\text{p}^g \mid \vect{\beta}_\text{p}^g]
$
denote the group-wise scale and transformation matrix that align the group-relative point maps $\bm{X}^{i,g}$ to the point maps $\bm{X}^i$ expressed in the global reference frame. 
$\vect{\sigma}_{uv}^{i,g}$ denotes the uncertainty of the point map for frame $i$ in group $g$ at pixel $(u,v)$.
We further parameterize each of these point maps as
$
\bm{X}_{uv}^i
=
\mathbf{R}_\text{p}^{i^{\top}} 
\mathbf{K}_\text{p}^{i^{-1}} 
{\bm{D}^i}^{-1}_{\text{p}, uv}(u,v,1) 
+
\bm{o}_\text{p}^i
$
in terms of each camera's calibration $\mathbf{K}^i_\text{p}$, world-to-camera rotation $\mathbf{R}^i_\text{p}$, and center $\bm{o}^i_\text{p}$ expressed in the global reference frame, and the disparity map $\bm{D}^i_\text{p}$.
Substituting this expression into the loss function~\eqref{eq:align_point} and minimizing it, we can thus recover
$\mathbf{K}^i_\text{p}$,
$\mathbf{R}^i_\text{p}$,
$\bm{o}^i_\text{p}$,
$\bm{D}^i_\text{p}$,
$\lambda_\text{p}^g$,
$\bm{P}_\text{p}^g$
from the predicted point maps.

The steps above infer the disparity maps $\bm{D}_\text{p}^i$ from the point maps, but the model also predicts disparity maps $\bm{D}_\text{d}^i$ directly, where the $\text{d}$ subscript denotes disparity prediction.
We introduce the following loss to align them:
\begin{equation}
\label{eq:align_depth}
\mathcal{L}_{\text{d}}
\left(\bm{D}_\text{p}, \lambda_\text{d}^g, \beta_\text{d}^g\right)
=
\sum_{g \in \mathcal{G}}
\sum_{i \in g}
\left \|
\bm{D}_\text{p}^i
-\lambda_\text{d}^g
\bm{D}_d^{i,g}
-
\beta_\text{d}^g\right \|_1,
\end{equation}
where $\lambda_\text{d}^g$ and $\beta_\text{d}^g$ are optimized scale and shift parameters.

Finally, the ray maps $\bm{r}$ also encode camera pose.
To align them with the global camera parameters
$
(\mathbf{R}_\text{p}, \mathbf{K}_\text{p}, \bm{o}_\text{p})
$
obtained from the point map, we first solve an optimization problem to extract the camera parameters from the ray map
$\bm{r}^{i,g} = \langle
\bm{d}^{i,g},
\bm{m}^{i,g}
\rangle$
for each group $g$ at frame $i$.
Following Ray~Diffusion~\cite{zhangcameras},
the camera center
$\bm{o}^{i,g}_\text{c}$ is solved by finding the 3D world coordinate closest to the intersection of all rays:
\begin{equation}
\label{eq:opt_cam1}
\bm{o}^{i,g}_\text{c}
=
\underset{\bm{p}
\in
\mathbb{R}^3}
{\arg \min }
\sum_{u\in H, v\in W}\|
\bm{p}\times\bm{d}_{uv}^{i,g}
-
\bm{m}_{uv}^{i,g}\|^2 .
\end{equation}
The camera extrinsics are solved by optimizing for the matrix
$\vect{H}$ that transforms the predicted per-pixel ray directions
$\bm{d}_{uv}^{i,g}$ to the ray directions
$\mathbf{u}_{uv}$
of a canonical camera:
\begin{equation}
\label{eq:opt_cam2}
\vect{H}^{i,g}
=
\underset{\|\vect{H}\|=1}{\arg \min }
\sum_{u\in H, v\in W}
\left \|\vect{H}
\bm{d}_{uv}^{i,g}
\times
\mathbf{u}_{uv}
\right \|.
\end{equation}
Then the world-to-camera rotation matrix
$\vect{R}_c^{i,g}$
and intrinsic matrix
$\vect{K}_c^{i,g}$ can be solved using the RQ-decomposition of $\vect{H}^{i,g}$.
Finally, the camera trajectory alignment loss is:
\begin{align}
\label{eq:align_cam}
\mathcal{L}_\text{c}
(
\mathbf{R}_\text{p},
\bm{o}_\text{p},
\mathbf{R}_\text{c}^g,
\bm{\beta}_\text{c}^g,
\lambda_\text{c}^g)
=
\sum_{g \in \mathcal{G}}
\sum_{i \in g}
\Big(\left \|\mathbf{R}_\text{p}^{i^{\top}}
\mathbf{R}^{g}_\text{c}
\mathbf{R}^{i,g}_\text{c}
-\bm{I}
\right \|_{\mathrm{f}} \nonumber \\ 
+
\left \|
\lambda_\text{c}^g
\bm{o}_\text{c}^{i,g}
+\bm{\beta}_\text{c}^g
-\bm{o}_\text{p}^i\right \|_2 \Big), 
\end{align}
where $\bm{R}^{g}_\text{c},\bm{\beta}^{g}_\text{c}, \lambda^g_\text{c}$ are learnable group-wise rotation matrix, translation vector, and scale, respectively, to align the global camera trajectory
$(\mathbf{R}_p,\bm{o}_p)$
and the predicted ones
$(\mathbf{R}_c,\bm{o}_c)$.
Following MonST3R~\cite{zhang24monst3r}, we also use a loss to smooth the camera trajectory:
\begin{equation}
\label{eq:align_smooth}
\mathcal{L}_{\text{s}}
(\mathbf{R}_\text{p},
\bm{o}_p)
=
\sum_{i=1}^{N}
\left(\left \|
\mathbf{R}_\text{p}^{i^{\top}}
\mathbf{R}_\text{p}^{i+1}-\bm{I}
\right \|_{\mathrm{f}}
+\left \|\bm{o}_\text{p}^{i+1}
-\bm{o}_\text{p}^i\right \|_2\right).
\end{equation}
The final optimization objective is the weighted combination of the losses above:
\begin{equation}
\label{eq:opt_all}
\mathcal{L}_{\text {all}}
=
\alpha_1
\mathcal{L}_{\text {p}}
+ \alpha_2 \mathcal{L}_{\text {d}}
+ \alpha_3 \mathcal{L}_{\text {c}}
+ \alpha_4 \mathcal{L}_{\text {s}}.
\end{equation}

\tableVideoDepth

\paragraph{A note on the invariants.}

The model predicts point maps, disparity maps, and ray map origins up to scale, as this cannot be uniquely determined from a monocular video.
The disparity map is also recovered up to a translation, which discounts the focal length (this is sometimes difficult to estimate due to the dolly zoom effect).
Likewise, the ray map origin is recovered up to a shift, necessary to allow normalizing these maps.

\section{Experiments}%
\label{sec:exp}

\subsection{Experimental Settings}%
\label{eq:exp_setting}

\paragraph{Training datasets.}

\shortname is trained exclusively on synthetic datasets, yet demonstrates strong generalization to real-world videos.
Specifically, we use five synthetic datasets for training:
Spring~\cite{Mehl2023_Spring},
BEDLAM~\cite{black2023bedlamsyntheticdatasetbodies},
PointOdyssey~\cite{zheng2023pointodyssey},
TarTanAir~\cite{Wang2020TartanAirAD},
and VirtualKitti~\cite{cabon2020vkitti2}.
See the \suppmat~\cref{tab:ablation_Dataset} for details.

\paragraph{Training.}

Our \shortname is initialized with the weights of DynamiCrafter~\cite{xing2024dynamicrafter} and trained using AdamW~\cite{loshchilov2019decoupled} with a learning rate of $1 \times 10^{-5}$ and a batch size of 32.
We use a progressive training strategy to improve convergence and stability.
First, we train the model to generate a single geometric modality, \ie, the point maps, at a fixed resolution of $512 \times 320$.
Next, we introduce a multi-resolution training scheme to improve generalization and robustness, which includes various resolutions:
$512 \times 384$,
$512 \times 320$,
$576 \times 256$,
$640 \times 192$.
Finally, we progressively add additional geometric modalities, \ie, the ray and depth maps.
Training is conducted on 4 NVIDIA H100 GPUs with a total training time of approximately one week.

\paragraph{Inference.}

As described in~\cref{sec:method_4ddiffusion}, given an $N$-frame video as input, we first split it into overlapping clips $\mathcal{G}$, each containing $V=16$ frames, with a stride of $s=4$.
Each video clip is encoded and fed to the diffusion model to sample multi-modal 4D parameters $(\bm{X}^{i,g}, \bm{D}^{i,g}, \bm{r}^{i,g})$ for the video.
For sampling, we use DDIM~\cite{song2022denoisingdiffusionimplicitmodels} with 5 steps.
Finally, the alignment algorithm in~\cref{sec:method_4ddiffusion} is used to fuse the clips into a globally coherent 4D reconstruction of the entire video.

\subsection{Video Depth Estimation}%
\label{eq:exp_vde}

\paragraph{Testing data.}

Our hypothesis is that, despite being trained on synthetic data, our model can generalize well to out-of-distribution synthetic \emph{and} real data, as it is based on a pre-trained video diffusion model.
To test this hypothesis, we evaluate our model on three benchmarks:
Sintel~\cite{Butler2012ANO} is a synthetic dataset that provides accurate depth annotations, covering diverse scenes with complex camera motion.
KITTI~\cite{geiger13vision} is a large driving dataset collected using stereo cameras and LiDAR sensors.
Bonn~\cite{Palazzolo2019ReFusion3R} focuses on dynamic indoor scenes.
To ensure fair comparisons, we follow the evaluation protocol used by MonST3R~\cite{zhang24monst3r}, where depth sequences are uniformly sampled from the datasets, extracting 50--110 frames per sequence for evaluation.

\paragraph{Metrics.}

Following the standard affine-invariant depth evaluation protocol~\cite{Ranftl2019TowardsRM}, we align the predicted video depth with the ground-truth depth before computing metrics.
However, unlike single-image depth estimation~\cite{yang24depth,lihe24DAV2,ke2024repurposing}, where depth alignment is performed per frame, we enforce \emph{global scale consistency} by applying a single scale and shift across the entire video sequence.
For quantitative evaluation, we adopt two widely used depth metrics: absolute relative error (Abs Rel) and the percentage of inlier points (with a threshold value of $\delta < 1.25$).

\paragraph{Baselines.}

We compare \shortname to state-of-the-art
single-frame depth estimation methods (Marigold~\cite{ke2024repurposing} and Depth-Anything-V2~\cite{lihe24DAV2}),
video depth prediction (NVDS~\cite{wang23NVDS}, ChronoDepth~\cite{shao24learning}, and DepthCrafter~\cite{hu25depthcrafter}),
and joint video depth and camera pose prediction
(Robust-CVD~\cite{Kopf2020RobustCV}, CausalSAM~\cite{Zhang2022StructureAM}, and MonST3R~\cite{zhang24monst3r}).

\begin{figure*}[t]
    \centering
    \includegraphics[width=\textwidth]{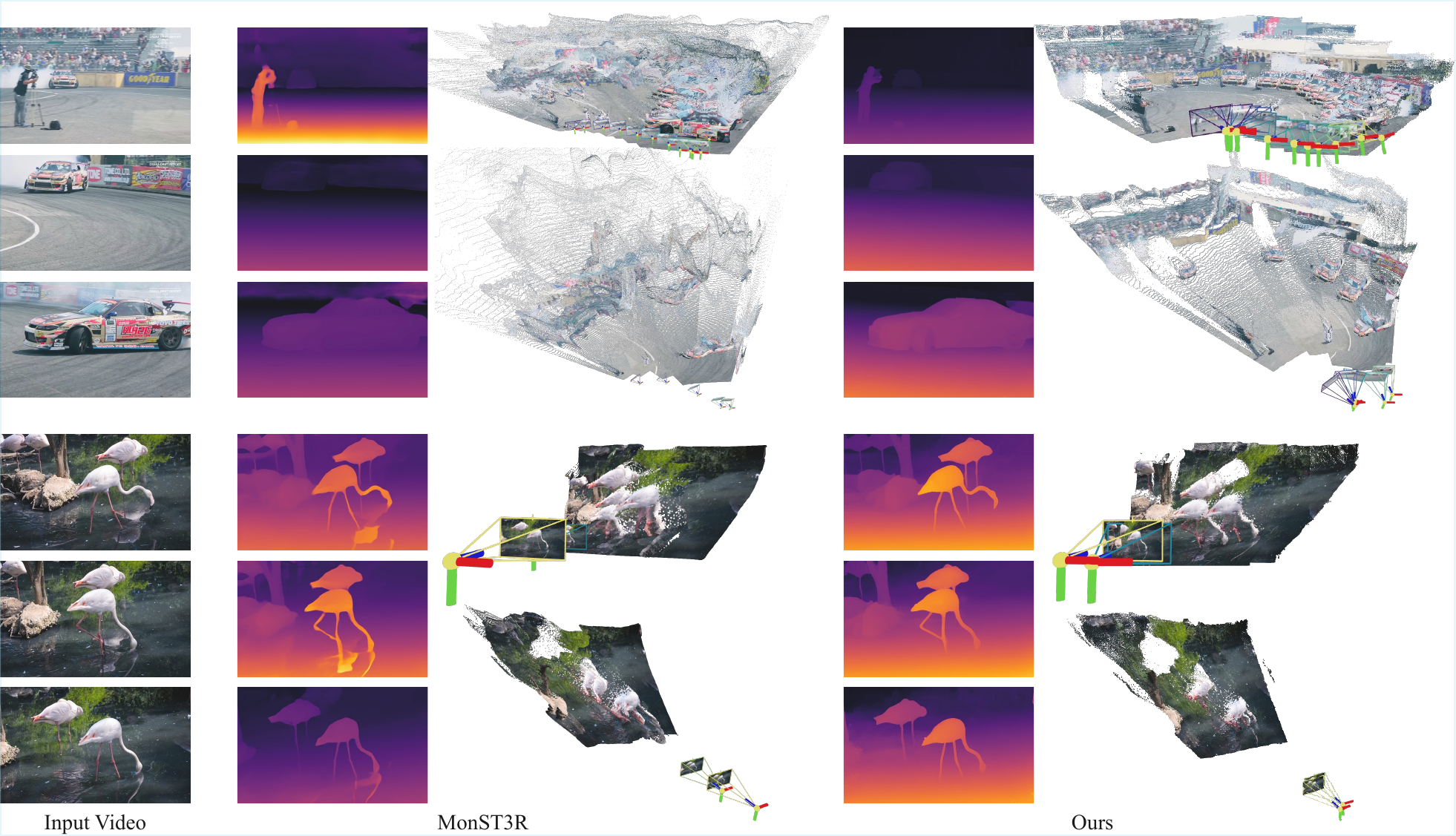}
     \vspace{-0.6cm}
    \caption{\textbf{Qualitative results} comparing \shortname with MonST3R~\cite{zhang24monst3r}. Attributed to our group-wise inference manner and prior geometry knowledge from pretrained video diffusion, our model successfully produces consistent 4D geometry under fast motion (first row) and deceptive reflection in the water (second row). }
    \label{fig:compare}
\vspace{-0.4cm}
\end{figure*}

\paragraph{Results.}

As shown in~\Cref{tab:video_depth}, all versions of \shortname outperform state-of-the-art methods by a large margin.
This includes DepthCrafter~\cite{hu25depthcrafter} and MonST3R~\cite{zhang24monst3r}, the most recent video depth diffusion model and the dynamic extension of DUSt3R to dynamic scenes, respectively.
Notably, while both \shortname and DepthCrafter are based on the same video diffusion model (DynamiCrafter), our model outperforms DepthCrafter in Abs Rel by $24.0\%$ on Sintel and $17.3\%$ on KITTI, despite solving a more general problem.
Qualitatively, \cref{fig:compare} shows that \shortname achieves more consistent results, especially for fast-moving objects.

\subsection{Camera Pose Estimation}%
\label{sec:exp_cam}

\paragraph{Setup.}

We evaluate the performance of \shortname on both the synthetic Sintel~\cite{Butler2012ANO} dataset and the realistic TUM-dynamics~\cite{Sturm2012ABF} dataset.
We follow the same evaluation protocol as in MonST3R~\cite{zhang24monst3r}.
Specifically, on Sintel, we select 14 dynamic sequences, and for TUM-dynamics, we sample the first 90 frames of each sequence with a temporal stride of 3.
After aligning the predicted camera trajectory with the ground truth using the Umeyama algorithm, we calculate three commonly used metrics:
Absolute Translation Error (ATE),
Relative Translation Error (RPE-T),
and Relative Rotation Error (RPE-R).
We compare our method with other state-of-the-art discriminative methods, which jointly predict camera pose and depth, including Robust-CVD~\cite{Kopf2020RobustCV}, CausalSAM~\cite{Zhang2022StructureAM}, and MonST3R~\cite{zhang24monst3r}.

\paragraph{Results.}

To the best of our knowledge, \shortname is the first method that uses a generative model to estimate camera parameters in a dynamic scene.
As shown in~\cref{tab:camera}, compared to existing non-generative alternatives, we achieve much better camera rotation prediction (RPE-R) and comparable camera translation prediction (ATE and RPE-T).

\begin{figure*}[t]
    \centering
    \includegraphics[width=\textwidth]{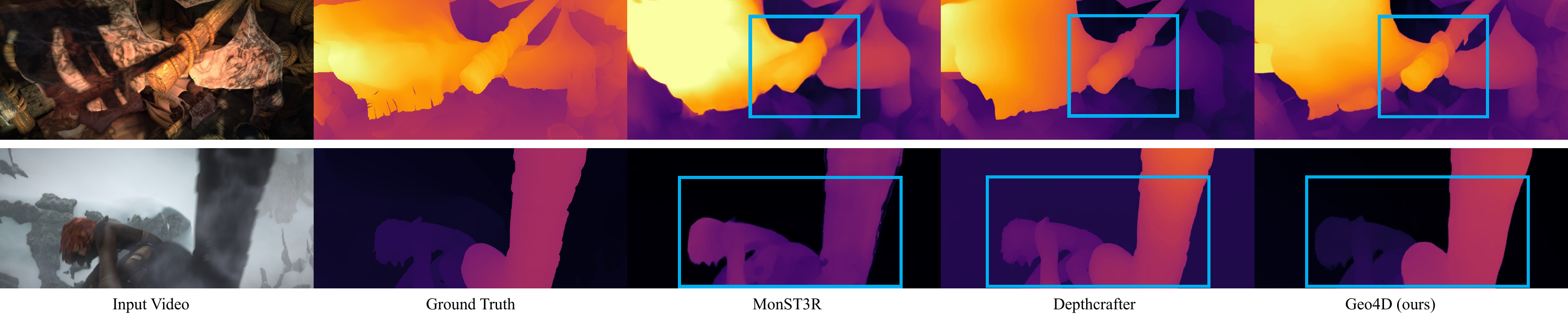}
     \vspace{-0.8cm}
    \caption{\textbf{Qualitative video depth results} comparing \shortname with MonST3R~\cite{zhang24monst3r} and DepthCrafter~\cite{hu25depthcrafter}.
    Owing to our proposed multi-modal training and alignment, as well as the prior knowledge from diffusion, our method can infer a more detailed structure (first row) and a more accurate spatial arrangement from video (second row).}%
    \label{fig:compare_depth}
\vspace{-0.4cm}
\end{figure*}

\tableCamerapose
\tableAblation

\subsection{Qualitative Comparison}

\paragraph{4D reconstruction.}

We compare \shortname with the state-of-the-art MonST3R method on the DAVIS~\cite{Huang_2016} dataset.
Upgrading from pairwise alignment as in MonST3R to our group-wise alignment improves temporal consistency, leading to a more stable and globally coherent 4D reconstruction of point maps and camera trajectory, particularly in highly dynamic scenes.
As shown in the top row of~\cref{fig:compare}, \shortname successfully tracks the racing car in 4D, whereas MonST3R struggles due to the rapid motion between pairs of images.
Furthermore, likely due to the strong prior captured by the pre-trained video generative model, \shortname correctly reconstructs the reflection of the flamingo in the water (second row in~\cref{fig:compare}), whereas MonST3R misinterprets the reflection as a foreground object, resulting in incorrect depth.

\tableStride

\paragraph{Video depth prediction.}

We compare \shortname with state-of-the-art video depth predictors MonST3R~\cite{zhang24monst3r} and DepthCrafter~\cite{hu25depthcrafter} on the Sintel~\cite{Butler2012ANO} dataset.
Qualitatively, \shortname produces more detailed geometry, for instance for the rope on the stick in the first row of \cref{fig:compare_depth}, and a better spatial arrangement between different dynamic objects, as shown in the second row of \cref{fig:compare_depth}.

\subsection{Ablation Study}%
\label{sec:exp_abl}

We ablate our key design choices and the effect of different modalities on the Sintel dataset.

We study the effect of multi-modality in \cref{tab:ablation}.
The three modalities---point map, disparity map, and ray map---can be used either at training or inference time, or both.
The first two rows show that the diffusion model trained with point maps as a single modality performs worse in both video depth and camera pose estimation than the diffusion model trained with all three modalities.
Therefore, the other two modalities, even if they can be seen as redundant, serve as additional supervisory signals during training, which improves the generalization ability of the diffusion model.

We then investigate the effectiveness of our multi-modal alignment algorithm.
Compared with the second to the fourth row in \cref{tab:ablation}, which leverage only a single modality during inference, multi-modal alignment optimization (last row) achieves the best performance, showing the benefits of fusing the multiple modalities at inference time.

We ablate the sliding window stride in \cref{tab:ablation_S}.
Results improve with a shorter stride, in part because this means that more windows and estimates are averaged, reducing the variance of the predictions by the denoising diffusion model, which is stochastic.
We choose stride $s=4$ for our main results to balance runtime and performance.
Note that MonST3R~\cite{zhang24monst3r} requires 2.41 seconds to process one frame under the same setting, so our method is 1.27 times faster than MonST3R~\cite{zhang24monst3r}.

\section{Discussion and Conclusion}%
\label{sec:con}

We have introduced \shortname, a novel approach that adapts a video generator for dynamic 4D reconstruction.
By building on a pre-trained video generator, \shortname achieves excellent generalization to real data despite being trained only on synthetic 4D data.
We have also demonstrated the benefits of predicting multiple modalities and fusing them at test time via optimization.
Our model outperforms state-of-the-art methods on video depth and camera rotation prediction, particularly in challenging dynamic scenes.

Despite these successes, our approach has limitations.
One is that the point map encoder-decoder is still not entirely accurate, which in turn is a bottleneck for the overall reconstruction quality.

Our approach also opens a path to integrating 4D geometry into video foundation models, \eg, to generate 3D animations from text, or to provide a more actionable signal when the video model is used as a proxy for a world model.

\paragraph*{Acknowledgments.}

The authors of this work were supported by Clarendon Scholarship, ERC 101001212-UNION, and EPSRC EP/Z001811/1 SYN3D.

{
    \small
    \bibliographystyle{ieeenat_fullname}
    \bibliography{main,chuanxia_general,chuanxia_specific,vedaldi_general,vedaldi_specific}
}

\clearpage
\maketitlesupplementary

In this \textbf{supplementary material}, we provide additional information to supplement our main submission. 
The \textbf{code} is available here for research purposes: \rurl{github.com/jzr99/Geo4D}

\section{Implementation Details}

\subsection{Training Dataset}

As shown in \cref{tab:ablation_Dataset}, we use five synthetic datasets for training:
Spring~\cite{Mehl2023_Spring},
BEDLAM~\cite{black2023bedlamsyntheticdatasetbodies},
PointOdyssey~\cite{zheng2023pointodyssey},
TarTanAir~\cite{Wang2020TartanAirAD},
and VirtualKitti~\cite{cabon2020vkitti2}.
Although all datasets are synthetic, we found that some depth pixels are missing in PointOdyssey~\cite{zheng2023pointodyssey}.
To address this, we apply max pooling to inpaint the missing pixels.
During training, we sample each dataset according to the ratios in \cref{tab:ablation_Dataset}.
For each sample, we select 16 frames from the sequence, with the sampling stride randomly chosen from $\{1,2,3\}$ to allow our diffusion model to adapt to input videos with various frame rates.

\subsection{Optimization Details}

The overall optimization process is outlined in~\cref{alg:A}.
We first predict all three modality maps using our diffusion model for each video clip $g$.
The predicted point maps are then roughly aligned based on the overlapping frames using the Umeyama algorithm~\cite{88573}.
The camera intrinsic $\textbf{K}^k$ is initialized by minimizing the projection error of the point map $\bm{X}^{k,g^k}$ in its reference (first) frame $k$ within each window group $g^k$.
The camera extrinsics are then initialized using the RANSAC PnP algorithm.
In the first stage of optimization, the point maps are roughly disentangled into camera pose and depth map.
The disparity map is then aligned with the global depth inferred from point maps by solving~\cref{eq:align_depth} from the main paper to obtain the scale and shift parameters.
The camera parameters extracted from the predicted ray map are aligned with the global camera trajectory based on the reference (first) frame of each video clip $g$ via~\cref{eq:align_cam} from the main paper.
After initializing all the alignment learnable parameters, including rotation $\vect{R}^g_{*}$, scale $\lambda^g_*$, and shift $\vect{\beta}^g_*$ across different modalities, where $* \in \{\text{p},\text{d},\text{c}\}$, we jointly optimize all the learnable parameters by~\cref{eq:opt_all}.
Specifically, we set the weights for each loss term in~\cref{eq:opt_all} as $\alpha_1 = 1, \alpha_2 = 2, \alpha_3 = 0.005, \alpha_4 = 0.015$ to roughly equalize the scale of the different losses.

\algopt

\section{Additional Analysis}
\subsection{Ablating the Number of Denoising Steps}

\tableDataset
\tableStep

We study the influence of the number of denoising steps during inference.
As shown in \cref{tab:ablation_step}, the model achieves optimal performance after around 5 steps.
Compared to the video generation task, where a larger number of denoising steps usually produces a more detailed generated video, 4D reconstruction is a more deterministic task, which requires fewer steps.
Similar phenomena are also observed in~\cite{hu25depthcrafter}, which uses a video generator for video depth estimation.  

\begin{figure*}[t]
    \centering
    \includegraphics[width=\textwidth]{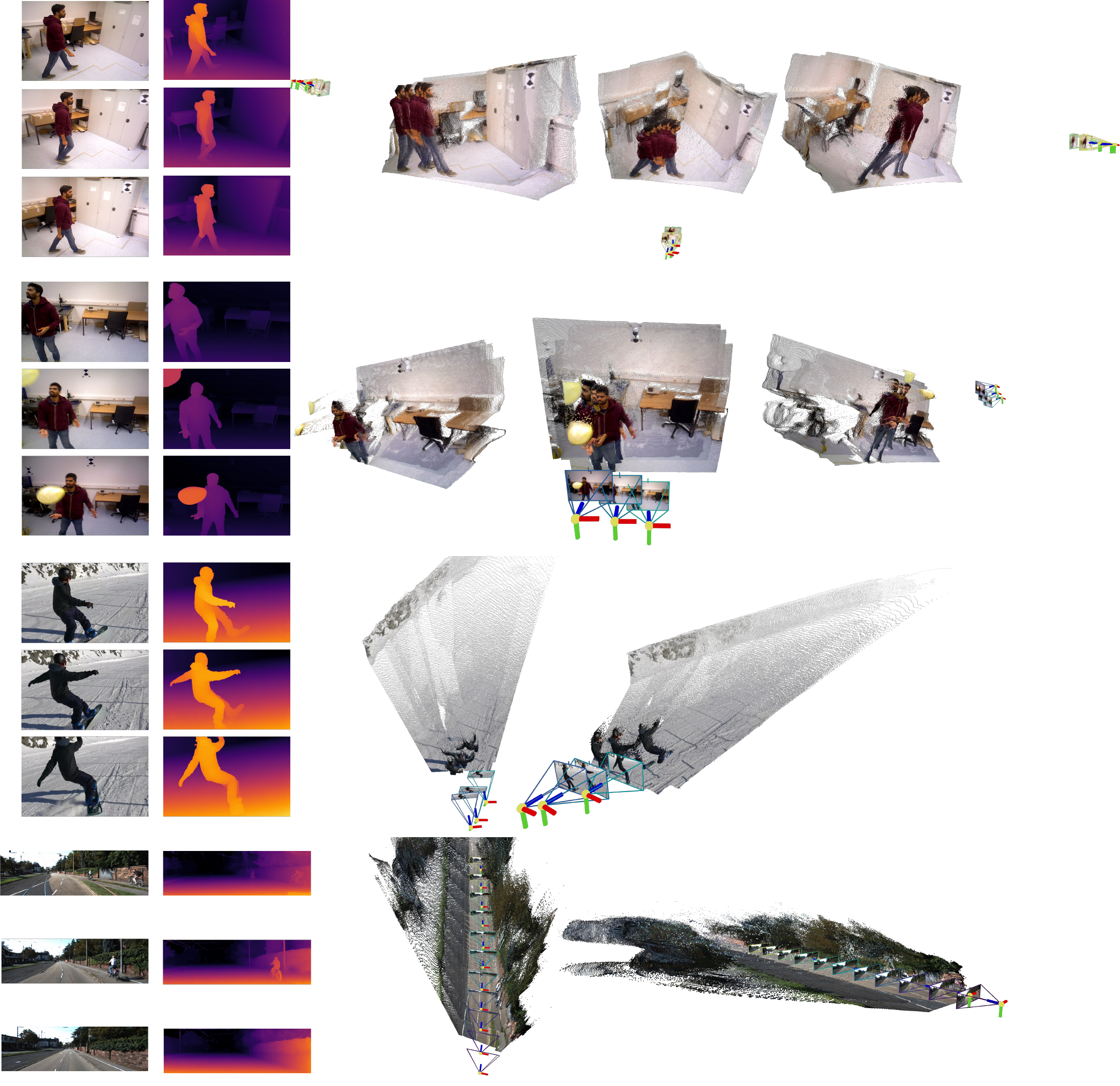}
    \caption{\textbf{Additional qualitative results.} Our method generalizes well to various scenes with different 4D objects and performs robustly against different camera and object motions.}%
    \label{fig:ad_compare}
\end{figure*}

\subsection{Ablation Study for Fine-Tuned Point Map VAE}

\tableVAE

As stated in the main paper, we added an additional branch to predict the uncertainty for our point map VAE and fine-tuned it based on Eq.~\ref{eq:loss_vae_point}.
We perform an ablation study on our fine-tuning strategy.
As shown in \cref{tab:ablation_VAE}, our fine-tuned point map VAE achieves consistently better performance on both video depth estimation and camera pose estimation tasks compared with the original pre-trained image VAE, demonstrating the necessity and effectiveness of our fine-tuning strategy.

\subsection{Analysis of Multi-Modal Representation}

\begin{figure}[t]
\centering
\includegraphics[width=\linewidth]{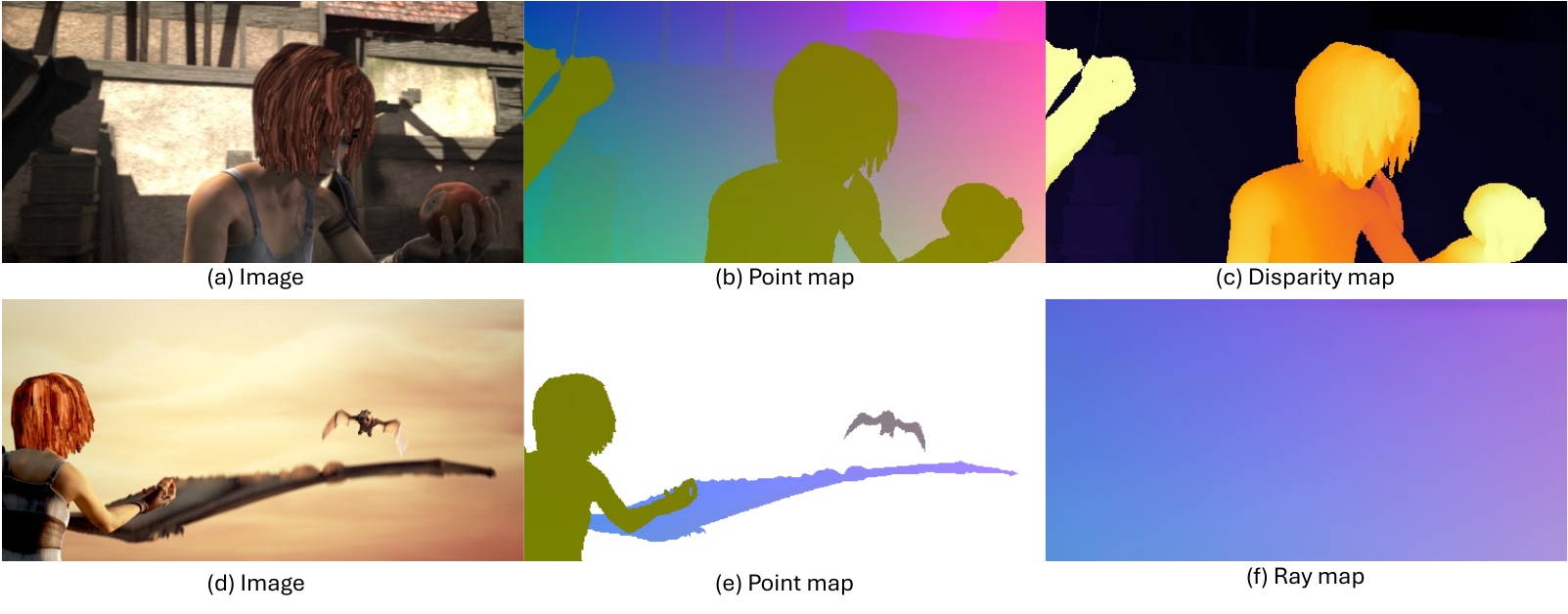}
\vspace{-0.5cm}
\caption{\textbf{Visualization of different geometric modality maps.}}%
\label{fig:modality}
\vspace{-0.3cm}
\end{figure}

Point maps (PMs) and disparity maps (DMs) are complementary.
DMs better represent near objects, while PMs are more depth-agnostic (\eg, human vs house in \cref{fig:modality}~(b,c)).
As in prior work, DMs are affine invariant (which here makes them range-compatible with the pretrained RGB VAE); their scale and shift, needed to recover undistorted geometry, are inferred by matching them to the predicted PMs.
Ray maps (RMs) help infer the camera pose when PMs fail to represent points at infinity (such as the sky in \cref{fig:modality}~(e)).
We observed that PMs tend to be noisier than DMs, so we prioritized modeling the PMs' uncertainty.
Per-pixel uncertainty for ray maps are less meaningful given the high degree of correlation between individual rays.
During multi-modal alignment, we align global point clouds with DMs in disparity space and with PMs in linear space. This naturally gives more weight to near points, which tend to be estimated well by DMs, and weighs points based on uncertainty with PMs, thus taking advantage of both modalities.

\section{Visualization}

\Cref{fig:ad_compare} shows additional visualizations for indoor, outdoor, and driving scenes.
Although our model is only trained on synthetic datasets, it generalizes to real-world data with diverse objects and motions.

\section{Limitations}

Although our method performs well and generalizes to a wide range of in-the-wild videos, it can struggle in cases involving significant changes in focal length or extreme camera motion throughout a sequence.
This limitation likely stems from the lack of focal length variation in our training data.
Incorporating more sequences with diverse camera movements and zooming effects could help mitigate this issue.
Moreover, due to the inherent temporal attention mechanism in our network architecture, our approach currently supports only monocular video input.
Extending the method to handle multi-view images or videos is a promising direction for future work.

\end{document}